\documentclass[sigconf,authorversion,nonacm]{acmart}
\pdfoutput=1
\AtBeginDocument{  }

\usepackage{soul} 
\usepackage[utf8]{inputenc}
\usepackage{amsmath}
\usepackage{booktabs}
\usepackage{tikz}
\usetikzlibrary{shapes,arrows,positioning,fit,backgrounds,matrix,chains,patterns,trees,3d}

\usepackage{drmacros}
\usepackage{multirow}
\usepackage{pgfplots}\pgfplotsset{compat=1.18}
\usepackage{tabularray}
\usepackage{colortbl}

\newcommand{\parai}[1]{\medskip\noindent{\bf #1}}

\newcommand{\para}[1]{\medskip\noindent{\bf #1:}}

\usepackage[many]{tcolorbox}
\usepackage{xcolor}

\newcommand{\fon}[1]{\fontfamily{#1}\selectfont}

\newtcolorbox{qbox}{colback=gray!10!white,
	colframe=gray!75!black,halign=center,drop shadow=black,left=1pt,right=1pt,top=1pt,bottom=1pt}
\AtBeginEnvironment{qbox}{\large}

\newtcolorbox{pbox}{colback=blue!5!white,
	colframe=blue!75!black,halign=justify,drop shadow=black,fontupper=\fon{ppl},box align=base,left=1pt,right=1pt,top=1pt,bottom=1pt,width=7.5cm}

\newtcolorbox{gtbox}{colback=green!5!white,
	colframe=green!75!black,halign=justify,drop shadow=black,fontupper=\fon{ppl},box align=base,left=1pt,right=1pt,top=1pt,bottom=1pt,width=7.5cm}

\newtcolorbox{gfbox}{colback=violet!5!white,
	colframe=violet!75!black,halign=justify,drop shadow=black,fontupper=\fon{ppl},box align=base,left=1pt,right=1pt,top=1pt,bottom=1pt,width=7.5cm}

\AtBeginEnvironment{qbox}{\large}

\newtcolorbox{cbox}[4]{colback=#4,
	colframe=#4,halign=center,drop shadow=black,fontupper=\fon{ppl},left=1pt,right=1pt,top=1pt,bottom=1pt,width=#2,title=#1,coltitle=#3}

\newcommand{\good}[1]{$+\hspace{0.14em}$ #1}
\newcommand{\bad}[1]{$-$ #1}

\newcommand{\colname}[1]{\multicolumn{1}{c}{\textbf{#1}}}
\newcommand{\catname}[1]{\textbf{\textsc{#1}}}
\newcommand{\scatname}[1]{\textsc{#1}}

\newcommand{\gbt}[1]{\begin{tabular}{@{}l@{}} #1 \end{tabular}}

\setcopyright{acmlicensed}
\copyrightyear{2025}
\acmYear{2025}
\acmDOI{XXXXXXX.XXXXXXX}

\acmConference[XXXX '25]{Make sure to enter the correct
	conference title from your rights confirmation emai}{}{}
\acmISBN{}

\begin{document}
	
	\title[Large Language Models, Knowledge Graphs and Search Engines: A Crossroads for Answering Users' Questions]{Large Language Models, Knowledge Graphs and Search Engines: A Crossroads for Answering Users' Questions}
	
	\author{Aidan Hogan}
	\email{ahogan@dcc.uchile.cl}
	\orcid{0000-0001-9482-1982}
	\affiliation{	\institution{IMFD; DCC, Universidad de Chile}
		\city{Santiago}
		\country{Chile}
	}
	
	\author{Xin Luna Dong}
	\email{lunadong@meta.com}
	\orcid{0009-0001-2049-2458}
	\affiliation{  \institution{Meta Reality Labs}
		\city{Redmond}
		\state{Washington}
		\country{USA}
	}
	
	\author{Denny Vrande\v{c}i\'c}
	\email{denny@wikimedia.org}
	\orcid{0000-0002-9593-2294}
	\affiliation{	\institution{Wikimedia Foundation}
		\city{San Francisco}
		\state{California}
		\country{USA}
	}
	
	\author{Gerhard Weikum}
	\email{weikum@mpi-inf.mpg.de}
	\orcid{0000-0003-4959-6098}
	\affiliation{	\institution{Max Planck Institute for Informatics}
		\city{Saarbrücken}
		\country{Germany}
	}
	
	\renewcommand{\shortauthors}{Hogan et al.}
	
	\begin{abstract}
		Much has been discussed 
		about how Large Language Models, Knowledge Graphs and Search Engines can be combined 
		in a synergistic manner.
		A dimension largely absent from current academic discourse is the user perspective. In particular, there remain many open questions regarding how best to address the diverse information needs of users, incorporating varying facets and levels of difficulty.
		This paper introduces a taxonomy of user information needs, which guides us to study the pros, cons and possible synergies of Large Language Models, Knowledge Graphs and Search Engines.
		From this study, we derive a roadmap for future research.
	\end{abstract}
	
	\newcommand{\ld}[1]{\textcolor{red!50!black}{\textbf{Luna:} #1}}
	\newcommand{\ah}[1]{\textcolor{orange!50!black}{\textbf{Aidan:} #1}}
	\newcommand{\dv}[1]{\textcolor{blue!50!black}{\textbf{Denny:} #1}}
	\newcommand{\gw}[1]{\textcolor{green!50!black}{\textbf{Gerhard:} \tt #1}}
	\newcommand{\LD}[1]{\ld{#1}}
	\newcommand{\AH}[1]{\ah{\bf\tt #1}}
	\newcommand{\DV}[1]{\dv{#1}}
	\newcommand{\GW}[1]{\gw{\bf\tt #1}}
	\newcommand{\gwins}[1]{\textcolor{green!50!black}{{#1}}}
	\newcommand{\gwdel}[1]{\st{#1}}
	
	\newcommand{\ahins}[1]{\textcolor{blue!70!black}{{#1}}}
	\newcommand{\ahdel}[1]{\textcolor{orange!50!black}{\st{#1}}}
	
	\newcommand{\ahli}[1]{\textcolor{blue!50!black}{{#1}}}
	\newcommand{\ahld}[1]{\textcolor{blue!50!black}{\st{#1}}}

	
	
	\received{XXX}
	\received[revised]{XXX}
	\received[accepted]{XXX}
	
	\maketitle
	
	\section{Introduction}
	
	Generative AI has led to heated debate on the opportunities, limitations and risks of such technology, and the precedent it sets for the future.
	These AI models and tools are often
	powered by \textit{Large Language Models} (\textit{LLMs}) 
	with many billions of parameters trained on vast corpora of natural language sourced 
	from the Web. 
	State-of-the-art LLMs are mostly based on the Transformer architecture~\cite{DBLP:conf/nips/VaswaniSPUJGKP17} for neural networks, trained to predict the next word or \textit{token} of characters for a given input text, which can be an incomplete sentence, a question or a task description, such as asking for travel advice, solving a puzzle, or writing code for a programming exercise. The models operate in an ``auto-regressive'' way, generating entire output sequences token by token. LLMs trained on broad corpora and for a wide spectrum of tasks are sometimes
	called \textit{foundational models}~\cite{DBLP:journals/corr/abs-2108-07258} as they show proficiency across domains and tasks: from mass-user chatbots to co-pilots for software developers.
	
	Deployed LLMs are often invoked with \textit{prompt engineering} or \textit{in-context learning} in order to best shape the application inputs towards desired outputs.
	Notable LLMs that are widely used include
	the OpenAI family of GPT models, Google's PaLM
	models, Huawei's PanGu series, Meta's Llama family, and more. 
	LLM-based assistants such as Google Gemini or Bing Co-Pilot
	are gaining popularity. 
	
	\begin{figure}
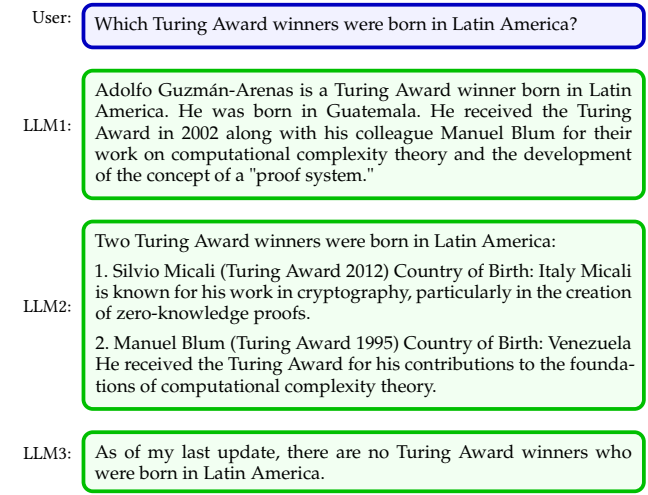
\footnotesize
		\begin{tblr}{@{}Q[t,r]@{\quad$\!\!$}Q[m,l]@{}}
			User: & \begin{pbox}Which Turing Award winners were born in Latin America?\end{pbox} \\[-5ex]
			LLM1: & \begin{gtbox}
				Adolfo Guzmán-Arenas is a Turing Award winner born in Latin America. He was born in Guatemala. He received the Turing Award in 2002 along with his colleague Manuel Blum for their work on computational complexity theory and the development of the concept of a "proof system."
			\end{gtbox} \\[-5ex]
			LLM2: & \begin{gtbox}
				Two Turing Award winners were born in Latin America:\\[1ex]
				1. Silvio Micali (Turing Award 2012)
				Country of Birth: Italy
				Micali is known for his work in cryptography, particularly in the creation of zero-knowledge proofs.\\[1ex]
				2. Manuel Blum (Turing Award 1995)
				Country of Birth: Venezuela
				He received the Turing Award for his contributions to the foundations of computational complexity theory.
			\end{gtbox} \\[-5ex]
			LLM3: & \begin{gtbox}As of my last update, 
				there are no Turing Award winners who were born in Latin America.\end{gtbox} \\
		\end{tblr}
		\caption{
			Replies from three LLMs to a user question \label{fig:chatgpt}}
	\Description[Replies of three LLMs to a user question]{Replies of three LLMs to a user question about Turing Award winners born in Latin America}		
	\end{figure}
	
	Notwithstanding these technical achievements and impressive applications, LLMs still have a variety of fundamental shortcomings.
	Figure \ref{fig:chatgpt} illustrates some of these, by showing the responses of three different state-of-the-art models to a user question about Turing laureates born in Latin America.
	LLM1 lists two winners, but the first never won the Turing award.
	LLM2 lists two people who are indeed both Turing laureates, but the first one is born in Italy, not Latin America.
	Finally, the last LLM gives an empty list with the proviso that it was trained a while ago. 
	The ground-truth is that only Manuel Blum should be in a correct reply.
	
\begin{figure}
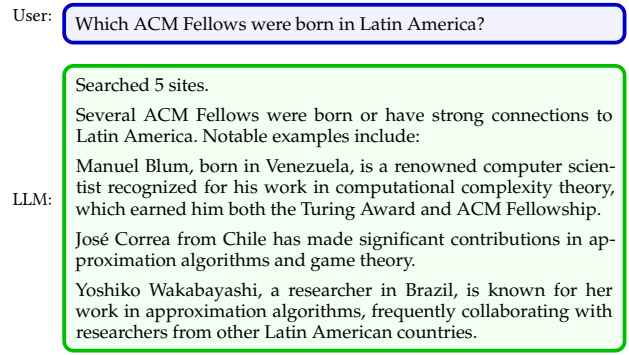
\footnotesize
		\begin{tblr}{@{}Q[t,r]@{\quad$\!\!$}Q[m,l]@{}}
			User: & \begin{pbox}Which ACM Fellows were born in Latin America?\end{pbox} \\[-5ex]
			LLM: & \begin{gtbox}
				Searched 5 sites.\\[1ex]
				Several ACM Fellows were born or have strong connections to Latin America. Notable examples include:\\[1ex]
				Manuel Blum, born in Venezuela, is a renowned computer scientist recognized for his work in computational complexity theory, which earned him both the Turing Award and ACM Fellowship.\\[1ex]
				José Correa from Chile has made significant contributions in 
				approximation algorithms and game theory.\\[1ex]
				Yoshiko Wakabayashi, a researcher in Brazil, is known for her work in approximation algorithms, frequently collaborating with researchers from other Latin American countries.
			\end{gtbox} \\[-5ex]
		\end{tblr}
		\caption{LLM reply to a long-tail question \label{fig:chatgpt-longtail-example}}
		\Description[Reply of one LLM to a long-tail question]{Reply of one LLM to a long-tail question about ACM fellows born in Latin America.}	
	\end{figure}

	While Turing laureates are prominent entities, other information requests refer to lesser-known ``long-tail'' topics.
	Figure \ref{fig:chatgpt-longtail-example} is a long-tail example, where the user asks for ACM Fellows from Latin America. The LLM response lists three people, none of whom is in the official list of ACM Fellows. The LLM reply even states that it ``\textit{searched 5 sites}'', so it made use of fresh Web content (via Retrieval Augmented Generation (RAG), discussed later). The ground-truth is found on the ACM web site\footnote{See \url{https://awards.acm.org/fellows/award-recipients?year=&award=158&region=Latin+America}; retr.\ \today.}, which currently lists 4 Fellows. When shifting the question to {\em ACM Fellows from Europe}, the LLM returns five names, based on ``searching 6 sites''. All of these are correct, but when insisting for a complete list, no further names are given. The ACM web site currently lists 201 European Fellows.\footnote{See \url{https://awards.acm.org/fellows/award-recipients?year=&award=158&region=Europe&submit=Submit&isSpecialCategory=}; retr.\ \today.}
	
	These scenarios, and many other observations discussed in the literature, highlight several fundamental limitations of LLMs:
	
	\begin{enumerate}
		\item \textit{Hallucination}: LLMs are vulnerable to ``inventing'' situations, facts, events, etc., that are not based on reality and the underlying data.
		\item \textit{Opaqueness}: It is often unclear what sources are involved in an answer, and how they were combined.
		\item \textit{Staleness:} Due to the computational and energy costs of training LLMs, 
		it is prohibitive to keep them up to date; they are bound to lag behind on many topics.
		\item \textit{Incompleteness}: 
		By their probabilistic nature, LLMs give ``best-effort'' replies, but, in many cases, they are unable to enumerate complete lists of answers.
	\end{enumerate}
	
	What can be done about such limitations? An approach gaining traction is to combine LLMs with other complementary technologies. 
	Specifically, combining the Information Retrieval (IR) techniques that power search engines (SEs) with LLMs gives rise to the idea of \textit{Retrieval Augmented Generation} (\textit{RAG})
	\cite{DBLP:conf/nips/LewisPPPKGKLYR020,DBLP:conf/kdd/FanDNWLYCL24}.
	Given a prompt, SEs are used to find relevant documents of data excerpts (e.g., from Web tables) that can be passed to the LLM as additional inference-time context, often improving the answer quality. Despite such improvements, per the long-tail example of Figure~\ref{fig:chatgpt-longtail-example}, these limitations sometimes persist even with RAG enabled, due either to being unable to retrieve relevant information, or due to the information retrieved being represented in a manner not conducive to in-context learning (e.g., being given as an image, a table, etc.).
	
	While RAG combines paradigms for retrieving (SEs) and learning (LLMs) from unstructured data (i.e., text), another promising direction is to combine LLMs (and potentially RAG) with \textit{structured} forms of data and knowledge. In this latter realm, \textit{Knowledge Graphs (KGs)} have been gaining attention 
	\cite{DBLP:journals/csur/HoganBCdMGKGNNN21}. 
	KGs use graph-based representations in order to structure, integrate, query and reason about diverse collections of data and knowledge. Starting with the Google Knowledge Graph in 2012, KGs have been adopted by a wide range of other companies in a variety of industries, such as Accenture, AirBnb, Amazon, Bloomberg, eBay, LinkedIn, Meta, Uber, and many more,
	with applications like semantic search, recommendations, fraud detection, personal agents, risk assessment, etc.
	Aside from enterprises, Open Knowledge Graphs, such as 
	BabelNet ({\small\url{https://babelnet.org/}}) \cite{DBLP:journals/ai/NavigliP12},
	DBpedia ({\small\url{https://dbpedia.org/}}) \cite{DBLP:conf/semweb/AuerBKLCI07}, Wikidata ({\small\url{https://wikidata.org/}}) \cite{DBLP:journals/cacm/VrandecicK14}, 
	and YAGO ({\small\url{https://yago-knowledge.org/}}) \cite{DBLP:conf/www/SuchanekKW07},
	are freely available online.
	
	Some 
	literature (e.g., Suri et al.~\cite{abs-2301-03980}) suggested, perhaps provocatively, 
	that LLMs 
	will render explicit KGs obsolete:
	if machines can 
	understand
	all digital content on the Internet,
	why should 
	we put effort into structuring data and knowledge? 
	From the previous examples, and as we will argue herein, this view is 
	not justified. 
	Rather the growing consensus -- and a growing body of research -- indicates that LLMs and KGs complement each other
	\cite{DBLP:journals/corr/abs-2306-08302,DBLP:journals/corr/abs-2306-11489,DBLP:journals/corr/abs-2308-10168,DBLP:journals/tgdk/PanRKSCDJO0LBMB23}.
	
	This paper offers an exploration of the strengths, limitations and potential synergies of SEs, KGs and LLMs from the perspective of 
	an information-seeking user.
	Specifically, we first present an analysis of the fundamental strengths and limitations of these technologies, followed by a categorization of different information needs, and how they can (or cannot) be addressed by these three technologies individually, or in unison. This culminates in a roadmap for research on combining these 
	assets to better serve users' needs.
	
	\para{Background} In order to keep the discussion self-contained, Appendix~\ref{sec:background} provides a concise background on SEs, KGs and LLMs, explaining the key techniques involved. A reader unfamiliar with one or more of these technologies is recommended to read through this appendix before continuing.
	
	\section{SE vs.\ KG vs.\ LLM}
	
	In this section we compare the strengths and limitations of SEs, KGs and LLMs. Similar analyses 
	have been presented
	for KGs and LLMs \cite{DBLP:journals/tgdk/PanRKSCDJO0LBMB23,DBLP:journals/corr/abs-2306-08302}; here we introduce SEs and a suite of new issues in our comparison. 
	Table~\ref{tbl:proscons} provides an overview of the pros and cons of SEs, KGs and LLMs 
	along various dimensions. When discussing the strengths and limitations of LLMs, we consider LLMs in isolation. Combinations of technologies (e.g., RAG) will be discussed later.
	
	\begin{table}
		\footnotesize
		\caption{Pros and cons of SEs, KGs and LLMs 
			\label{tbl:proscons}}
		\setlength{\tabcolsep}{1.3pt}
		\centering
		\begin{tabular}{llll}
			
			\toprule
			\colname{Dimension} & \colname{SE} & \colname{KG} & \colname{LLM} \\
			
			\midrule
			
			\catname{Correctness} & \gbt{\good{stores corpus}\\\bad{noisy content}} & \gbt{\good{stores corpus}\\\good{precise operators}} & \gbt{\bad{abstracts corpus}\\\bad{hallucinations}} \\ 
			\midrule
			
			\catname{Coverage} & \gbt{\good{broad coverage}} & \gbt{\good{domain specific}\\\bad{patchy coverage}} & \gbt{\good{broad coverage}\\\bad{poor long tail}} \\ 
			\midrule
			
			\catname{Completeness} & \gbt{\good{complete results}\\\bad{unclear scope}} & \gbt{\good{complete answers}\\\bad{unclear scope}} & \gbt{\bad{missing results}\\\bad{unclear scope}} \\ 
			\midrule
			
			\catname{Freshness} & \gbt{\good{quick updates}\\\good{news often text}} & \gbt{\good{deprecation}\\\bad{structure lag}} & \gbt{\bad{slow updates}\\\bad{cold start}} \\
			\midrule
			
			\catname{Generation} & \gbt{\bad{no generation}} & \gbt{\good{ontologies/rules}\\\good{graph learning}} & \gbt{\good{text generation}} \\ 
			\midrule
			
			\catname{Synthesis} & \gbt{\bad{no integration}\\\bad{no synthesis}} & \gbt{\good{data integration}\\\good{synthesis}} & \gbt{\good{text integration}\\\good{synthesis}} \\ 
			\midrule
			
			\catname{Transparency} & \gbt{\good{provenance}\\\bad{opaque ranks}} & \gbt{\good{algorithmic}\\\good{provenance}} & \gbt{\bad{black box}\\\bad{no provenance}} \\
			\midrule
			
			\catname{Coherency} & \gbt{\good{consistent} \\ \bad{rank variance}} & \gbt{\good{consistent} \\ \good{static analysis}} & \gbt{\bad{inconsistent} \\ \bad{(opt.) randomness}} \\
			\midrule
			
			\catname{Refinability} & \gbt{\good{refinable}\\\bad{opaque ranks}} & \gbt{\good{refinable}} & \gbt{\bad{indirect refinement}\\\bad{unpredictable}} \\ 
			\midrule
			
			\catname{Fairness} & \gbt{\good{content as-is}\\\bad{bias in ranks}} & \gbt{\good{content as-is}\\\bad{bias in coverage}} & \gbt{\bad{generative biases}\\\bad{needs safeguards}} \\
			\midrule
			
			\catname{Usability} & \gbt{\good{natural language}\\\good{simple queries}} & \gbt{\bad{structured}\\\bad{complex queries}} & \gbt{\good{natural language}\\\good{conversational}} \\
			\midrule
			
			\catname{Expressivity} & \gbt{\bad{simple queries}\\\bad{ambiguity}} & \gbt{\good{complex queries}\\\bad{lacks nuance}} & \gbt{\good{complex queries}\\\bad{ambiguity}} \\
			\midrule
			
			\catname{Efficiency} & \gbt{\good{retrieval-based}\\\good{simple queries}} & \gbt{\good{retrieval-based}\\\bad{complex queries}} & \gbt{\bad{inference-based}\\\bad{costly training}} \\
			\midrule
			
			\catname{Multilingualism} & \gbt{\bad{lang.\ dependent}} & \gbt{\good{lang.\ agnostic}\\\bad{manual labels}} & \gbt{\good{multilinguality}\\\bad{variable results}} \\
			\midrule
			
			\catname{Personalization} & \gbt{\bad{limited context} \\ \bad{not interactive}} & \gbt{\bad{limited context}\\
				\bad{not personalizable}} & \gbt{\good{in-context learn.}\\
				\good{personalizable}} \\
			\midrule
			
		\end{tabular}
	\end{table}

	\parai{Correctness} refers to the extent to which the information returned is correct. SEs and KGs 
	store explicit representations of their (indexed) contents: full text documents, tables and nodes/edges. LLMs, on the other hand, only capture statistical patterns from the input corpus that can be used to generate text following patterns similar to the input. 
	These patterns are stored in latent form, encoded in the values of billions of model parameters.
	Thus while imprecise answers generated by SEs and KGs can usually be traced back to issues with the input corpus, LLMs may generate imprecise results, hallucinations, etc., not supported at all by their input data. 
	As a consequence, LLMs can perform poorly regarding topics with few supporting documents (i.e., long-tail topics)~\cite{DBLP:conf/acl/MallenAZDKH23}.
	
	\parai{Coverage} refers to the broadness and comprehensiveness of the content covered. With respect to the coverage of the raw information on which they are based, SEs index vast collections of documents crawled from the Web. LLMs are likewise often trained on vast collections of text, including the content of Web crawls, as well as selected complete corpora of particular human interest, such as Wikipedia, news archives, book collections, etc., but may not retain long-tail facts seen few times in the corpus. KGs, on the other hand, are repositories of curated data in a structured representation. They cover many millions of entities and billions of facts about them: this is very high coverage, but still only a tiny fraction of what is textually and noisily stated in the corpora of the largest SEs and LLMs. SEs and LLMs dominate in terms of the broadness of their coverage, which is delimited by the wider domain of information available in natural language, but there exist domain-specific KGs for verticals like finance, health, energy and more, which can be key assets for applications where data quality is crucial.
	
	\parai{Completeness} relates to the ability to return all relevant information from the base corpus. KGs are capable of returning a complete set of answers with respect to their content. SEs are also capable of generating a complete (though paginated) list of matching results from their indexes, though these may not represent a complete set of answers for a user's question. LLMs will often generate incomplete answers, even with respect to the corpus they were trained on (per Figure~\ref{fig:chatgpt-longtail-example}). For example, although music catalogs, personal websites and social media list many thousands of musicians who have covered Bob Dylan songs, it is virtually impossible to recall a near-complete list using SEs or LLMs alike. However, in the case of all three technologies, it is often unclear when results are (in)complete. 
	
	\parai{Freshness} captures the degree to which the information returned is up-to-date. LLMs are costly to train and are only re-computed and released occasionally, limiting their usefulness when fresh data is needed about current events~\cite{DBLP:journals/corr/abs-2310-03214}. Furthermore, more recent events generally have less online content with which to train LLMs, further deepening the ``cold start'' problem. 
	SEs have update policies that prioritize news feeds and social media; so this content is captured in near-real-time. Finally, KGs can be continuously updated, and out-of-date information can be explicitly deprecated. KGs can be equally up-to-date or more up-to-date than natural language sources when tapping into live structured data streams for events, relating for example to finance, lotteries, sports, weather, etc. However, where such streams are not available, or not used, the quality standards imposed by KGs may require additional curation, verification or expertise that creates an update lag versus other sources. For example, while the Wikidata KG has tens of thousands of users active in the past month, this pales in comparison to the number of users posting about current events on social media, meaning that Wikidata can lag behind, particularly for long-tail events.
	
	\parai{Generation} encapsulates the ability to derive custom, novel content from the base corpus in order to better address a user's information need. LLMs are generative by design, allowing text to be recursively generated from a starting prompt. KGs are not generative \textit{per se}, though ontologies and rules can generate new knowledge via deductive reasoning, while knowledge graph embeddings and graph neural networks can generate new knowledge via inductive reasoning; techniques such as summarization are also generative in nature. SEs, 
	by themselves,
	are also not generative, and thus can only return base information to users largely ``as is''.
	
	\parai{Synthesis} references the ability to return individual results that integrate information from multiple sources (addressing what some refer to as ``global queries''). KGs and LLMs are capable of returning answers synthesized from multiple sources. While SEs index diverse sources, they cannot synthesize information to answer user queries; thus, if the information a user seeks is not provided in one document, the user must integrate and synthesize the relevant information manually from multiple sources.
	
	\parai{Transparency} speaks of the ability to reproduce or otherwise understand how and why information is returned with respect to the base corpus. KGs are transparent, where the results provided for a query can be explained 
	in terms of their derivation steps and (where available) underlying data provenance.
	SEs are also transparent, though some aspects, like ranking algorithms, may be more opaque. LLMs, as black-box models, are largely opaque, meaning that it is difficult to understand how or why they generate particular responses, or what sources were used.
	
	\parai{Coherency} refers to the ability to produce logically-consistent responses across repetitions of the same request or invocations of logically-related requests over the same base corpus. Repeating the same request to SEs and KGs will typically yield deterministic results, with changes in output linked to changes in the input corpus (though in the case of SEs, ranking methods may show high variance for small changes). Conversely, LLMs are often (and SEs are sometimes) made non-deterministic by design, which enables more diverse -- but less reproducible -- responses. While LLMs can be configured to be more or less deterministic, the responses may not be coherent for seemingly equivalent requests. For example, we cannot know \textit{statically} (looking just at the requests) whether or not an LLM will return coherent responses for ``\textit{countries that are islands}'' and ``\textit{islands that are countries}''~\cite{DBLP:journals/tacl/JiangXAN20} even though both requests are semantically equivalent. For an analogous request over KGs and SEs, we can prove \textit{query equivalence}, i.e., we can assert that two requests will return the same response over any dataset, even if, for example, conjuncts (like \textit{island}/\textit{country}) are reordered (though in the case of SEs, rankings may vary). We can also, for example, prove \textit{query containment} for KGs and SEs, such that the responses for ``\textit{countries that are islands}'' are contained in the responses for ``\textit{countries}'' or for ``\textit{islands}'', which is not feasible for LLMs.
	
	\parai{Refinability} speaks of the ability to apply edits to improve the quality of information returned in future. Incomplete or incorrect results from a KG or an SE are likely due to missing or erroneous input, which can be addressed by a user with write privileges. Refining LLMs in order to remove an erroneous result or address an incomplete answer is more indirect at best and infeasible at worst. 
	The indirect way is to identify the troublesome text(s) in the training corpora, edit or remove them, and re-run the entire training and fine-tuning process. 
	However, there is no guarantee that even this will fix the issue, as the questionable output may stem from multiple pieces mashed together in the generative process.
	Without re-training the LLM, fixing errors is non-trivial. Despite some works pursuing parameter-weight editing for the latent model, there is no principled way of ensuring the desired effects.
	
	\parai{Fairness} refers to returning unbiased information. SEs, KGs and LLMs can all suffer from bias. While SEs and KGs present (potentially biased) content ``as is'', the situation is more troublesome for LLMs since they generate text. LLMs can then regurgitate and even amplify biases, prejudices and other harmful tropes found in their training corpora~\cite{DBLP:conf/ci2/KotekDS23}.  While safeguards and 
	mitigation steps can be applied 
	to avoid generating objectionable content, such measures sometimes block legitimate studies of sensitive topics. While SEs are not generative, the content they index and return may be biased, and ranking measures may also introduce bias. KGs can also contain biased facts, or can be biased in terms of coverage~\cite{DBLP:conf/www/Demartini19}, but this can be addressed over time via curation.
	
	\parai{Usability} references the ease with which users can request the information they need. SEs and LLMs are based on natural language, which greatly boosts their usability; they ``speak'' our language. Conversely, to exploit the full power of KGs, users must write structured queries 
	in a formal language: a high barrier for many.
	Thus, while users often interact directly with SEs and LLMs, users most often interact indirectly with KGs via end-user 
	applications.
	On the positive side, natural language interfaces that map user questions (e.g., ``\textit{Which Turing Award winners were born in Latin America?}'') into structured queries (e.g., Cypher, SPARQL, etc.) are being advanced to improve KG usability.
	
	\parai{Expressivity} refers to the ability to express and respond to potentially complex requests.
	KGs support expressive query languages that can join data across relations, group entities by categories such as geo-regions, compute aggregates (e.g., averaging) over groups, and much more
	\cite{DBLP:journals/csur/AnglesABHRV17}.
	LLMs and SEs can process arbitrary requests in natural language (with varying success). 
	For SEs, these are traditionally in the form of keywords, but telegraphic queries and full-fledged questions are equally supported today. However, the input length is tightly limited.
	LLMs, on the other hand, accept fairly long user text (e.g., 50K tokens and increasing), which allows natural conversations and even working with user-provided data such as spreadsheets.
	But operating over text has pros and cons.
	LLMs 
	can capture nuanced expressions that are difficult to model in KGs.
	On the other hand,  KGs feature formal types and logical operators, with unambiguous interpretations not possible for text.
	For example, negating a statement can be very subtle and tricky to understand in language, whereas it is explicit and clear in logic.
	
	\parai{Efficiency} refers to the low consumption of resources (time, memory, energy, etc.) for addressing a particular request. SEs and KGs index their content, allowing for efficient retrieval. In the case of KGs, more complex queries -- such as path queries, queries with many joins, etc. -- may require more resources to evaluate. Nevertheless, with a good query optimizer and a query of moderate complexity, the execution time is usually in the order of seconds (or less).
	LLMs process user inputs not by retrieval, but by inference, making even simple requests (e.g., ``\textit{What is the capital of Venezuela?}'')
	relatively costly to process in terms of time, energy and computational resources~\cite{DBLP:conf/hpec/SamsiZMLMJBKTG23}.
	Moreover, the inference utilizes a large GPU platform with massive parallelism, whereas most KG queries are handled on a single CPU core. Comparable user requests are thus orders of magnitude more expensive to address with a contemporary LLM in terms of computational and environmental costs.
	
	\parai{Multilingualism} captures the ability to seamlessly handle base information, requests and responses in different languages with comparable performance. SEs and LLMs can provide divergent responses for analogous requests in different languages. While LLMs do have multilingual support, low-resource languages often exhibit
	drops in performance~\cite{DBLP:conf/emnlp/ZhangLHSK23}, where results can vary across languages. Conversely, KGs apply a graph-based abstraction of data and knowledge that is largely agnostic to natural language, though nodes and edges need to be (often manually) associated with human-readable names and descriptions in different languages in order for the KG to be applicable in a multilingual scenario~\cite{DBLP:journals/tgdk/KaffeeBKVM23}.
	
	\parai{Personalization} speaks of the ability to adapt responses to better address the needs of a particular user in a particular context. LLMs are capable of 
	in-context learning, which enables high degrees of personalization of responses, and interactivity, whereby previous prompts and responses shape future responses. SEs and KGs typically leverage more limited implicit forms of context, 
	like spatio-temporal context, previous queries, etc., with limited personalizability.
	However, more interactive UIs are being developed for KGs
	and other structured data resources
	\cite{DBLP:journals/sigweb/LissandriniPHM20}, while interfaces also continue to improve for SEs and LLMs.

	\section{The Perspective of Information-Seeking Users}
	
	The previous section provides some general insights into the pros and cons of SEs, KGs and LLMs with respect to several key issues. The question we now ask is: \textit{how do these factors affect the end user?} The answer to this question clearly depends on what information need the user is trying to satisfy. In some cases, the user's need may be satisfied with SEs; in other cases an LLM may suffice; in yet other situations a KG may be useful; finally, there are also cases where none of these technologies currently suffices. Towards a more meaningful answer to this question, we first draft a conceptual list of user needs. Some such categories have been identified, for example in the creation of benchmarks~\cite{abs-2406-04744}, but previous categorizations tend to focus purely on factual questions, whereas our categorization covers a broader range of typical user questions. This list can then be compared against the capabilities (and limitations) of SEs, KG and LLMs. Table~\ref{tbl:needs} presents a categorization of common user needs along these lines that we now discuss.
	
	\begin{table*}
		\setlength{\tabcolsep}{3pt}
		\caption{Categorization of users' information needs \label{tbl:needs}}
		\footnotesize
		\begin{tabular}{llllll}
			\toprule
			\colname{Category} & \colname{Subcategory} & \colname{Examples} & \colname{SE} & \colname{KG} & \colname{LLM} \\
			
			\midrule
			
			\multirow{15}{*}{\catname{Facts}} & \scatname{Popular} & \gbt{\textit{Who directed the movie Spotlight?}\\
				\textit{Who was the 2023 MVP of the NFL?}} &  \gbt{\good{good coverage}\\\bad{noisy content}} & \gbt{\good{good correctness}\\\bad{poor coverage}} & \gbt{\good{good coverage}\\\bad{noisy content}} \\ \cmidrule{2-6}
			
			& \scatname{Long-Tail} & \gbt{\textit{Which galaxy is closest to the Sunflower Galaxy?}\\						\textit{In which sports does the Ferroportuária club compete?}
			} &  \gbt{\good{good coverage}\\\bad{needle in haystack}} & \gbt{\good{good correctness}\\\bad{sparse coverage}} & \gbt{\bad{limited storage}\\\bad{hallucinations}} \\ \cmidrule{2-6}
			
			& \scatname{Dynamic} & \gbt{ \textit{What is the stock price of Alphabet?}\\ 			\textit{When is the next match of the Delhi Capitals?} } &  \gbt{\good{fresh results}\\\good{live websites}} & \gbt{\bad{structure lag}\\\bad{poor coverage}} & \gbt{\bad{sporadic training}\\\bad{cold start}} \\ \cmidrule{2-6}
			
			& \scatname{Multi-hop} & \gbt{\textit{Which Turing Award winners were born in Latin America?} \\ 						\textit{Which La Liga players transferred from Asian clubs?}}
			& \gbt{\bad{no reasoning}\\\bad{single-shot search}} & \gbt{\good{formal reasoning}\\\good{structured queries}} & \gbt{\good{latent reasoning}\\\bad{hallucinations}} \\ \cmidrule{2-6}
			
			& \scatname{Analytical} & \gbt{\textit{How many U.S. Congress Members are younger than 50?} \\ 						\textit{Who scored the most 3-pointers in NBA finals?} } & \gbt{\bad{no datatypes}\\\bad{no aggregation}} & \gbt{\good{rich datatypes}\\\good{aggregation}} & \gbt{\bad{no datatypes}\\\bad{no aggregation}} \\ \midrule
			
			\multirow{6}{*}{\catname{Explanations}} & 
			\scatname{Commonsense} & \gbt{\textit{How do snakes move?} \\ \textit{What is the offside rule in football?} } & \gbt{\good{good coverage}\\\good{text output}} & \gbt{\bad{poor coverage}\\\bad{structured output}} & \gbt{\good{good coverage}\\\good{text output}} \\ \cmidrule{2-6} 
			
			& \scatname{Causal} & \gbt{			\textit{What caused the dancing plague of 1518?} \\ 			\textit{Why did New Zealand lose the 2023 Rugby World Cup final?}
			} & \gbt{\good{good coverage}\\\good{text output}} & \gbt{\good{long tail}\\\bad{structured output}} & \gbt{\good{good coverage}\\\good{text output}} \\ \cmidrule{2-6}
			
			& \scatname{Exploratory} & \gbt{ \textit{Who was Williamina Fleming?} \\ 			\textit{How is boxing linked with early-onset dementia?}}  & \gbt{\good{text output}\\\good{ranked results}} & \gbt{\good{graph algorithms}\\\good{browsing + navigation}} & \gbt{\good{interactive}\\\good{synthesis}} \\ \midrule
			
			\multirow{6}{*}{\catname{Planning}} & 
			\scatname{Instructive} & \gbt{\textit{How do I tie a Windsor Knot?} 			\\ \textit{How to organize a beach volleyball tournament?} } & \gbt{\good{multimedia}\\\good{diverse results}} & \gbt{\bad{poor coverage}\\\bad{non-didactic output}} & \gbt{\good{interactive}\\\bad{no multimedia}} \\ \cmidrule{2-6} 
			
			& \scatname{Recommendation} & \gbt{\textit{What are the best card games for Android?} \\ 			\textit{Which Winter Olympics events should I watch?}} & \gbt{\good{diverse results}\\\good{ranked results}} & \gbt{\bad{poor coverage}\\\bad{no recommendations}} & \gbt{\good{interactive}\\\good{synthesis}}  \\ \cmidrule{2-6} 
			
			& \scatname{Spatio-temporal} & \gbt{\textit{Which Thai restaurants near Disneyland open until midnight?}\\ 			\textit{What sport events are held in Glasgow in May?}
			} & \gbt{\good{events \& maps}\\\bad{no integration}} & \gbt{\good{integration}\\\good{s.-t. operators}} & \gbt{\bad{lacks freshness}\\\bad{no s.-t. operators}}  \\
			\midrule
			
			\multirow{6}{*}{\catname{Advice}} & 
			\scatname{Lifestyle} & \gbt{\textit{How can I improve my work/life balance?} \\ 			\textit{Is football more dangerous than basketball?}} & \gbt{\good{diverse results}\\\good{ranked results}} & \gbt{\bad{poor coverage}\\\bad{lacks nuance}} & \gbt{\good{interactive}\\\good{synthesis}} \\ \cmidrule{2-6}
			
			& \scatname{Cultural} & \gbt{\textit{Should I tip bartenders in Canada?} \\ 			\textit{Why do Arsenal and Chelsea fans hate each other?}} & \gbt{\good{diverse results}\\\good{ranked results}} & \gbt{\bad{poor coverage}\\\bad{lacks nuance}} & \gbt{\good{synthesis}\\\bad{cultural bias}} \\ \cmidrule{2-6} 
			
			& \scatname{Philosophical} & \gbt{
				\textit{As a vegan, is it okay to stop my kids eating meat?} \\ \textit{Could a stunning football goal be considered a work of art?} 			} & \gbt{\good{diverse results}\\\good{ranked results}} & \gbt{\bad{poor coverage}\\\bad{lacks nuance}} & \gbt{\good{interactive}\\\good{synthesis}} \\
			\bottomrule
		\end{tabular}
	\end{table*}
	
	\para{Facts} Users seek objective, verifiable information that may be satisfied by a simple answer: a name, number, list, table, etc. 
	
	\begin{description}
		\item[\normalfont\textsc{Popular}] queries seek simple factual information that is of broad interest, and thus frequently sought. Such queries are straightforwardly solved by SEs, KGs, and LLMs, though LLMs in particular may fail to address queries whose answers have only come to light after the training of the model, and indeed, may occasionally hallucinate and give an incorrect answer.
		
		\item[\normalfont\textsc{Long-Tail}] queries seek simple factual information that is of narrow interest, and thus infrequently sought. SEs and LLMs struggle with such queries, with few if any webpages providing explicit answers, thus also reducing the training data for LLMs. Where available, domain-specific KGs can help to address long-tail queries in that particular domain.
		
		\item[\normalfont\textsc{Dynamic}] queries have answers that vary in real-time, such as the stock price of a company, the score in an ongoing sports game, etc. SEs can often provide real-time information sourc\-ed from the Web for prominent events or domains, whereas KGs (if not exploiting a relevant data stream) and LLMs will often struggle due to lag in terms of structuring knowledge or training models, respectively.
		
		\item[\normalfont\textsc{Multi-hop}] queries require combining together multiple relations (for example, which people have won Turing Awards, which countries they were born in, and which countries are in Latin America) in order to provide an answer. SEs and LLMs again struggle with such queries for similar reasons to the long-tail, combined with their limited (deductive) reasoning capabilities. In this category, structured queries using joins and path expressions over KGs can excel, particularly if combined with deductive reasoning.
		
		\item[\normalfont\textsc{Analytical}] queries require computing novel -- typically numerical -- information from a given set of facts in order to derive an answer.\footnote{Note that we would consider a question such as ``\textit{how many cantons does Switzerland have?}'' to be a popular query, as its answer is likely to be explicitly available, and thus does not require computing potentially new information.} SEs and LLMs again struggle with such queries, which require numerical computations to derive potentially novel information~\cite{DBLP:conf/nips/HendrycksBKABTS21}. Furthermore, SEs and LLMs operate over text, which lacks formal datatypes. In this category, structured queries with aggregation and numerical functions over appropriate KGs with datatypes can excel.
	\end{description}
	
	\para{Explanations} Users seek an explanation as to what something is, what caused it, what properties it has, how it works, etc., based principally on objective criteria.
	
	\begin{description}
		\item[\normalfont\textsc{Commonsense}] queries request explanations of commonsense know\-ledge~\cite{DBLP:journals/cacm/DavisM15}, for example, why the sky is blue, how snakes move, etc. SEs and LLMs will excel here, as there will often be many explanations provided on the Web for this. LLMs have the additional advantage of being interactive, allowing users to quiz further about a particular explanation. KGs will be of limited use here, though they might yield relevant facts.
		
		\item[\normalfont\textsc{Causal}] queries involve explanations of what caused events or facts that may or may not be widely known. While SEs and LLMs can perform well here for more well-known phenomena, where causal relationships are complex and often more easily expressed in natural language, they may lack the information necessary to explain more obscure phenomena. Conversely, while KGs may capture causal relationships, even for obscure phenomena, typically what caused something is nuanced, and nuance is difficult to capture in a structured form.
		
		\item[\normalfont\textsc{Exploratory}] queries involve gaining initial understanding into an unfamiliar topic, or further understanding into a familiar one. In this case, the user may not know precisely what they are looking for, but will rather hope to recognize it when they see it. For these sorts of queries, SEs, LLMs and KGs have complementary strengths and weaknesses. SEs can be used to find a wealth of authoritative sources to explore a particular topic. LLMs can provide interactive dialog to shift to or narrow down on something of interest. KGs can also offer distinctive graph-based exploratory features, such as faceted browsing, path finding between topics or entities, entity similarity, graph summarization, etc.
	\end{description}
	
	\para{Planning} Users seek information to help them take a particular action or make a more informed decision. Such needs may involve a mix of objective and subjective criteria.
	
	\begin{description}
		\item[\normalfont\textsc{Instructive}] queries seek procedural knowledge about how to complete a particular task, often as a list of sub-tasks. SEs excel here, particularly for common tasks, as often they will lead to multimedia content on a website, such as a video or set of images, illustrating how to perform the task; they may also yield alternative webpages providing a variety of perspectives or didactic techniques that a user can benefit from. While LLMs can produce a list of instructions, again for common tasks, text outputs may not be the best way to illustrate procedures (though LLMs are increasingly incorporating multimedia features that could be useful in future for such questions). KGs typically do not focus on representing procedural knowledge; for example, while Wikidata contains items referring to a variety of national dishes, and what they contain, it provides no information on how to prepare them.
		
		\item[\normalfont\textsc{Recommendation}] queries seek general suggestions on a preferred course of action based on subjective and objective criteria. SEs and LLMs can again help when the topic is well-covered on the Web. SEs can provide diverse opinions and results, with ranking helping to highlight those most prominent or useful. The synthesis of LLMs helps to capture a consensus, and interactivity enables discussion about the topic, personal preferences, etc., in real time. KGs cannot directly address such questions, and have no inherent recommendation faculties, though their factual content may be useful to inform a particular recommendation, particularly for long-tail topics. 
		
		\item[\normalfont\textsc{Spatio-temporal}] queries seek information in a particular geographical or temporal setting, relating to, for example, events, places, times, etc. Again, the capabilities of SEs, LLMs and KGs are hit-and-miss for such queries. SEs can help find up-to-date information about particular events via the Web, and can lead users to help map information, but cannot help with drawing information together. LLMs may be able to help with some static geo-spatial information, or regarding historical events, and are further able to personalize results within a given context, but will fare poorly for long-tail places, contemporaneous events, etc. Many KGs contain rich geographic information that can be queried using geo-spatial operators, but they again may not be fully up-to-date, and be only indirectly helpful for assessing subjective criteria.
	\end{description}
	
	\para{Advice} Users seek general counsel that may range from matters of self-improvement to the ethical or philosophical. Such questions are often more open-ended than those covered in the other categories, and may again blur the lines of the subjective and the objective.
	
	\begin{description}
		\item[\normalfont\textsc{Lifestyle}] queries seek general advice about how users can improve their living situation, often involving a mix of subjective and objective criteria. Responses to such queries are typically not structured, with the best sources of information being prior responses and discussions for similar questions. In this setting, SEs and LLMs can perform well, with the former providing diverse results, and with the latter providing a synthesis reflecting a common consensus. Meanwhile KGs may only be of indirect help.
		
		\item[\normalfont\textsc{Cultural}] queries request guidance about cultural norms, etiquette, etc. SEs and LLMs can be well-suited to such tasks, particularly for common themes relating to widely-known cultures. Their textual output is again suitable for returning advice. However, the responses of LLMs, in particular, may be biased towards or against certain cultural perspectives~\cite{DBLP:conf/ci2/KotekDS23}. Again, KGs are not likely to be of much direct help, with the narrow exception perhaps of a KG dedicated to such cultural norms. The nuances of such queries are generally not compatible with structured responses.
		
		\item[\normalfont\textsc{Philosophical}] queries deal with more abstract topics of life, meaning and the world. They may not permit a crisp answer, but rather a discussion of different perspectives. In this sense, SEs and LLMs can again excel in this sub-category. While KGs may provide more factual or historical information regarding particular philosophies or philosophers, they would only be of indirect use.
	\end{description}
	
	\para{Other considerations} Across all query types, certain transversal issues may arise that require technical solutions, some of which are within the capabilities of SEs, KGs and LLMs, while others require external services or considerations.
	
	\begin{itemize}
		\item Questions may involve \textit{deixis}, including expressions such as “here”, “yesterday”, “me”, etc. that must be resolved in the user's context; as an example, an analytical query might ask ``\textit{How many U.S. Congress Members are younger than me?}". Such questions highlight challenges on how such expressions can be resolved while maintaining the privacy of users.
		\item Questions may involve \textit{multilingual information} in order to bridge from one language to another; as an example, a monolingual English speaker looking at a French menu might ask for a recommendation: ``\textit{What are some good seafood dishes?}''. The multilingual features of LLMs and KGs, in particular, should assist in such scenarios.
		\item Questions may involve \textit{multimedia information}, such as for the commonsense question: ``\textit{What animal is depicted in this picture?}''. This may require the use of multi-modal models that go beyond language, as are now being investigated.
		\item Questions may involve \textit{false premises}, i.e., assumptions that do not hold; for example, ``\textit{When did John F.\ Kennedy commit suicide?}''. In an SE setting, users may be biased to search for a particular answer, while LLMs often provide an answer even if the premise of the question itself is faulty~\cite{DBLP:journals/corr/abs-2310-03214}.
		\item Questions may involve \textit{ambiguity}, as in the case of the factual question ``\textit{Who is the mayor of Springfield?}'', where there are a great many places called Springfield. Such ambiguity may require interaction with the user to resolve, independently of the technology used.
		\item Questions may be the subject of \textit{controversy} or varying opinions, such as the question ``\textit{Would building more nuclear power plants help the environment?}''. Such questions may require a nuanced way to respond that reflects potentially divergent viewpoints.
	\end{itemize}
	
	Additionally, answers to questions may be imperfect due to incomplete, biased or incorrect information; thus, answers may require \textit{verification} or \textit{provenance}, to be useful. This is feasible for SEs and KGs, but much more complicated in the case of LLMs.
	
	\para{Summary} Our categorization reveals some trends regarding the strengths and weaknesses of SEs, KGs and LLMs for addressing different types of user needs. KGs help to deal with potentially-complex factual queries over curated knowledge in a precise, efficient and transparent manner; however, they are difficult to use, and do not help for more subjective criteria, nor for capturing the nuance required for explanations, planning or advice. SEs can help with all categories of queries, but only under the supposition that a document in the corpus exists that succinctly provides a (near-)direct answer; otherwise, if the requisite information is spread over multiple documents, combining it is left as an exercise to the user. LLMs help with all categories of queries, and can help to combine perspectives, facts, etc., from multiple documents, but can suffer from hallucinations, failing also to recognize false premises in questions; their performance for more obscure, domain-specific, multi-hop or analytical questions leaves room for improvement (per Figure~\ref{fig:chatgpt}), particularly  when combined with their lack of transparency.
	
	\section{Research Directions}
	
	The discussion so far has revealed that SEs, KGs and LLMs are complementary technologies: where one is weak, often another is strong.
	This motivates 
	exploring amalgams of these approaches that maximize their constituents' pros while minimizing their cons.
	
	Revisiting Table~\ref{tbl:proscons}, 
	for example, one can ask: what combination of SEs, KGs and LLMs can go beyond the individual technologies
	in supporting more categories of information needs?
	The ideal solutions should provide Internet-scale coverage and freshness, give precise answers with user-friendly explanations of provenance, and support the entire spectrum from fact lookups to analytic queries and personalized advice, all with low computational cost.
	
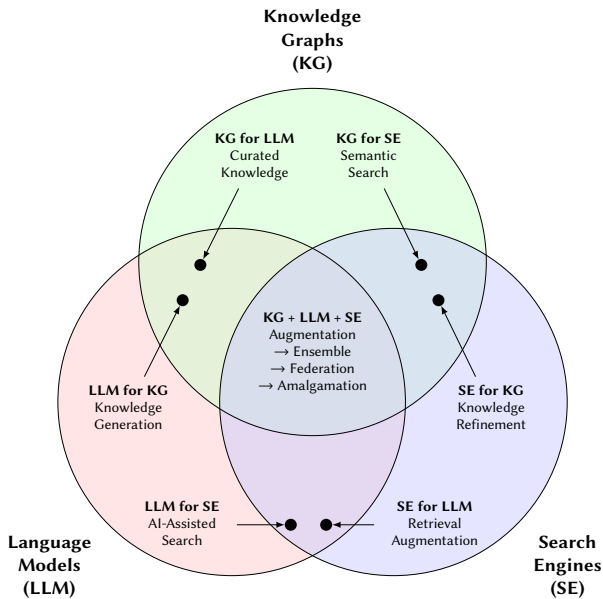
\begin{figure}
		\def\firstcircle{(0,0) circle (2.9cm)}
		\def\secondcircle{(60:2.7cm) circle (2.9cm)}
		\def\thirdcircle{(0:2.7cm) circle (2.9cm)}
		\resizebox{\columnwidth}{!}{
			\begin{tikzpicture}[
				every node/.style={
					font=\sffamily
				}
				]
				\begin{scope}[shift={(3cm,-5cm)}, fill opacity=0.5, text opacity=1]
					\fill[red!20] \firstcircle;
					\fill[green!20] \secondcircle;
					\fill[blue!20] \thirdcircle;
					
					\draw \firstcircle node (llm) [below left=2.9cm] {\textbf{\begin{tabular}{c}Language\\Models\\(LLM)\end{tabular}}};
					\draw \secondcircle node (kg) [above=3cm] {\textbf{\begin{tabular}{c}Knowledge\\Graphs\\(KG)\end{tabular}}};
					\draw \thirdcircle node (se) [below right=2.9cm] {\textbf{\begin{tabular}{c}Search\\Engines\\(SE)\end{tabular}}};
					
					\node (kag) at ($(kg)!0.7!(llm)$) {\footnotesize \begin{tabular}{c}\textbf{LLM for KG}\\Knowledge\\Generation\end{tabular}};
					
					\node (gak) at ($(llm)!0.78!(kg)$) {\footnotesize \begin{tabular}{c}\textbf{KG for LLM}\\Curated\\Knowledge\end{tabular}};
					
					\node (kar) at ($(kg)!0.7!(se)$) {\footnotesize \begin{tabular}{c}\textbf{SE for KG}\\Knowledge\\Refinement\end{tabular}};
					
					\node (rak) at ($(se)!0.78!(kg)$) {\footnotesize \begin{tabular}{c}\textbf{KG for SE}\\Semantic\\Search\end{tabular}};
					
					\node (rag) at ($(se.north west)!0.82!(llm.north east)$) {\footnotesize \begin{tabular}{c}\textbf{LLM for SE}\\
							AI-Assisted\\Search
					\end{tabular}};
					
					\node (gar) at ($(llm.north east)!0.8!(se.north west)$) {\footnotesize \begin{tabular}{c}\textbf{SE for LLM}\\
							Retrieval\\Augmentation
					\end{tabular}};
					
					\node[circle,minimum size=0.2 cm,inner sep=0pt, outer sep=0pt, fill=black, fill opacity=1] (kagp) at ($(kag)!0.43!(gak)$) {}
					edge[latex-, fill opacity=1] (kag);
					
					\node[circle,minimum size=0.2 cm,inner sep=0pt, outer sep=0pt, fill=black, fill opacity=1] (gakp) at ($(gak)!0.43!(kag)$) {}
					edge[latex-, fill opacity=1] (gak);
					
					\node[circle,minimum size=0.2 cm,inner sep=0pt, outer sep=0pt, fill=black, fill opacity=1] (karp) at ($(kar)!0.43!(rak)$) {}
					edge[latex-, fill opacity=1] (kar);
					
					\node[circle,minimum size=0.2 cm,inner sep=0pt, outer sep=0pt, fill=black, fill opacity=1] (rakp) at ($(rak)!0.43!(kar)$) {}
					edge[latex-, fill opacity=1] (rak);
					
					\node[circle,minimum size=0.2 cm,inner sep=0pt, outer sep=0pt, fill=black, fill opacity=1] (garp) at ($(gar)!0.43!(rag)$) {}
					edge[latex-, fill opacity=1] (gar);
					
					\node[circle,minimum size=0.2 cm,inner sep=0pt, outer sep=0pt, fill=black, fill opacity=1] (ragp) at ($(rag)!0.43!(gar)$) {}
					edge[latex-, fill opacity=1] (rag);
					
					\node[below = 3.7cm of kg] {\footnotesize \begin{tabular}{c}\textbf{KG + LLM + SE}\\Augmentation\\$\rightarrow$ Ensemble\\$\rightarrow$ 
							Federation \\$\rightarrow$ 
							Amalgamation
					\end{tabular}};

				\end{scope}
			\end{tikzpicture}
		}
		
		\caption{Potential combinations of SEs, KGs and LLMs
			\label{fig:venn}}
		\Description[Venn-like diagram]{Venn-like diagram showing how SEs, KGs and LLMs can complement each other.}	
	\end{figure}

	The idea of combining LLMs with other technologies is not new.
	Most notably, KGs have been leveraged by SEs for entity-centric queries, and recent LLMs are coupled with SE techniques for
	RAG.
	This is not a complete picture, though, and still
	far from the full potential of cross-technology synergies.
	Figure~\ref{fig:venn} depicts a Venn diagram that 
	highlights major directions towards potential synergies.
	The following subsections discuss pairwise combinations first, and then provide ideas on three-way integration.
	
	\subsection{Augmenting Language Models}
	
	\subsubsection*{KG for LLM: Curated Knowledge} Per the old idiom, {``\textit{just because it's said doesn't make it true}''}, much of the text on which LLMs are trained is not factual in nature. KGs as a curated source of structured knowledge can thus enhance LLMs in a variety of ways~\cite{DBLP:journals/tgdk/PanRKSCDJO0LBMB23,DBLP:journals/tkde/YangCLDW24}, particularly for addressing complex (e.g., multi-hop or aggregation-based) factual questions.
	This can involve 
	pre-training and fine-tuning enhancements, or inference-time injection of factual knowledge. 
	For training, the input data can be annotated or otherwise enriched with factual information from the KG (e.g., entity markup in text documents, injecting entity types etc.).
	Another approach is to train both the LLM parameters and KG embeddings in a coupled manner
	(see, e.g., \cite{DBLP:conf/nips/YasunagaBR0MLL22}). For inference, the input prompts can be augmented by adding background knowledge. 
	For example, when asking the LLM for ACM Fellows at US universities, it would be useful to inject triples from the KG, stating that ACM is an organization in the field of Computer Science and possibly even providing a list of US universities that large KGs should have.
	Yet another, simple but effective, approach could be to check LLM outputs against KGs as a post-processing step for veracity assessment.
	
	Research along all these lines can help to reduce hallucinations and bias in LLM responses -- particularly in the context of long-tail and multi-hop factual questions -- but must address the challenges posed by the patchy and incomplete nature of KGs, and the need to reliably disambiguate which KG entities are mentioned in a text.
	
	\subsubsection*{SE for LLM: Retrieval-Augmented Generation} A very active area of research, and one heavily adopted and refined in practice, relates to the use of search engine (i.e., information retrieval) technologies in order to enable Retrieval-Augmented Generation (RAG)~\cite{DBLP:conf/nips/LewisPPPKGKLYR020,DBLP:conf/aaai/0011LH024}. This offers great potential for overcoming problems with non-factuality, staleness and bias in LLM responses, improving performance in particular for dynamic and long-tail factual queries.
	In the simplest RAG setting,
	documents relevant to a prompt are searched for at inference run-time and their content is injected into the context window of the LLM to improve 
	the generated responses.
	Recent AI assistants from big providers call a major SE to this end, but it is equally useful to inject in-house information for vertical domains, such as finance or health.
	More recent approaches
	advocate for Active 
	RAG
	\cite{DBLP:conf/emnlp/JiangXGSLDYCN23}, where relevant documents are iteratively retrieved during the generation process as needed, particularly when generating long texts. 
	RAG will 
	continue to be a very lively area, with various benchmarks gaining major attention
	(e.g., \cite{DBLP:conf/aaai/0011LH024}).
	 
	Regarding 
	Tables~\ref{tbl:proscons}~\&~\ref{tbl:needs},
	{RAG helps to improve LLMs on several dimensions:
		freshness, fact awareness, better responses for long-tail queries, provenance with tangible sources, and more.}
	
	\subsection{Augmenting Search Engines}
	
	\subsubsection*{LLM for SE: AI-Assisted Search}
	LLMs have the capability to enrich SE functionality on both the user-input side and the way results are presented
	\cite{DBLP:journals/corr/abs-2308-07107}. 
	On the input side, LLMs provide users with natural-language dialogue that humans appreciate, particularly for explanations, planning and advice. Queries are seamlessly derived from the user's utterances, based on the LLM's skills in language generation, such as paraphrasing. 
	This setting also paves the way for conversational IR \cite{DBLP:journals/ftir/ZamaniTDR23}, where users interactively explore certain topics.
	The processing of search requests can be also supported by LLMs, leveraging their inherent language skills to help find more matches to search queries and compute better relevance scores for individual results. 
	On the output presentation, LLMs can generate more direct answers from top-ranked results or overview-style summaries of the topics that the user wants to explore. These may even come with suggestions to the user on how to proceed further, in a human-friendly conversational manner.
	
	With regard to
	Tables~\ref{tbl:proscons}~\&~\ref{tbl:needs}, 
	LLM-augmented SEs could bring
	precise answers for more complex queries
	into the realm of feasibility,
	most notably, for information needs on planning or advice, and also perhaps for analytical queries.
	Note, however, that efficiency may be a showstopper. More research is needed on how to balance the adequacy of responses and the necessary computational resources.
	
	\subsubsection*{KG for SE: Semantic Search} Major search engines have 
	deployed
	\textit{semantic search} functionalities
	by means of their back-end KGs, which allow for answering entity-centric queries with 
	KG excerpts, sometimes called ``knowledge panels'' or ``knowledge cards''. For example, a keyword query ``\textit{Manuel Blum}'' or ``\textit{Blum CMU}'' may return the KG entity along with notable facts. 
	Though SEs are increasingly leveraging KGs in order to improve search results, with Google's Knowledge Graph being a prominent example, there is much more that can be done in this space. For example, the results for multi-hop requests like ``\textit{Turing Award winners who work at US universities}'' on modern SEs may return some relevant or prominent examples of entities, but the list will be patchy in comparison with what could be found via a structured query over a KG such as Wikidata.
	Similar observations can be made about aggregation queries, such as ``\textit{number of Turing Award winners from Europe}'', with better answers available via KGs than SEs.
	Limitations in SE results can also be found in requests that feature \textit{both} facts and text from webpages.
	Consider the dynamic, multi-hop query
	``\textit{today's news stories about EU heads of government}''. Current SEs return documents that contain ``\textit{EU}'', ``\textit{European}'', ``\textit{leaders}'', but miss results about heads of government where those keywords are missing. 
	Such results could be improved by interpreting ``\textit{EU heads of government}'' as a type and asking the KG to enumerate 
	instances of this type, that is, relevant politicians for which news stories can then be retrieved. 
	
	With regard to Tables~\ref{tbl:proscons}~\&~\ref{tbl:needs}, advancing the use of KGs would help SEs to compute precise results to more complex queries, particularly multi-hop queries.
	Making this progress calls for enhancing the SEs' abilities for reliable entity recognition and disambiguation, and mapping phrases that denote types or predicates as well.
	
	\subsection{Augmenting Knowledge Graphs}
	
	\subsubsection*{LLM for KG: Knowledge Generation}
	
	The provocatively titled work ``\textit{Language Models as Knowledge Bases?}''~\cite{DBLP:conf/emnlp/PetroniRRLBWM19} initiated
	a wave of research 
	to investigate if and to what extent an
	LLM can generate a full-fledged KG, based on advanced prompt engineering and few-shot in-context learning, and sometimes even with supervised fine-tuning.
	Such techniques typically take as input a subject--predicate pair, such as Venezuela--capital, plus further prompting and context, and generate one or more objects that complete the desired fact. 
	Results show that LLMs can indeed generate large volumes of facts, with decent accuracy, though still exhibiting some hallucinations. 
	However, the size of these generated fact collections are three or more orders of magnitude smaller than what state-of-the-art KGs contain (e.g., 100 million entities, 1 billion facts in Wikidata), and the output correctness is substantially lower than the standard set by the best KGs (requiring 95\% or even 99\% correctness; see, e.g., \cite{DBLP:journals/corr/abs-2407-13578}). There is then the danger that the hallucinations of LLMs could lead to polluting KGs, dragging down their correctness, and the quality of responses to factual queries (in particular).
	Rather than trying to re-create KGs of this scale and quality, an interesting direction would be to identify gaps in the KG 
	-- in terms of missing entities and facts about entities --
	and use LLMs to strategically fill such gaps. 
	In particular, less prominent entities in the long tail would call for extending KGs, but studies so far show that the LLM ability to generate facts rapidly degrades for long-tail entities that appear more rarely in the training corpus~\cite{DBLP:conf/icml/KandpalDRWR23,Veseli:EMNLP2023}. Conversely, the simple structure of KGs that make them a flexible solution for data integration scenarios inhibits their ability to represent more subtle aspects of knowledge. For this reason, KGs in practice often adorn the core graph abstraction with additional elements to capture such nuances; for example, Wikidata uses qualifiers, ranks and references to add additional context, priority and provenance to statements forming edges in the KG, allowing to succinctly but loosely capture, for example, that Bill Clinton was more prominent as a politician than a saxophonist. LLMs could help to enhance the ability of KGs to represent and capture more nuanced aspects of knowledge.
	
	Regarding
	Tables \ref{tbl:proscons} and \ref{tbl:needs}, such research would broaden the coverage and expressivity of KGs, 
	expanding the scope of requests they can answer.
	As an example, recall the multi-hop query about winners of ACM awards (not just the Turing Award, but all official ACM awards).
	Even if such queries are not exactly mainstream from the view of a major SE or LLM stakeholder, there is value in supporting specialized analysts, journalists and data scientists.
	
	\subsubsection*{SE for KG: Knowledge Refinement}
	
	SEs can be used to refine the knowledge of KGs in various ways relating to updates, verification, provenance, negation, etc.
	For dynamic queries, many current events are first publicized via social media, news articles, etc., on the Web, wherein SEs can be used to incorporate novel information into the KG, and to deprecate obsolete facts (e.g., when a new president is inaugurated). 
	SEs 
	are also vital for checking and validating or invalidating statements
	in a KG 
	by finding supportive or contradictory evidence, boosting correctness.
	In the context of the patchy coverage of KGs, 
	SEs are the main source for content to run fact-centric information extraction on. Many open KGs tap into Wikipedia as a primary source (via either tools or human contributors), but for richer scope, tapping into a broader set of webpages is necessary. However, the state of the art in information extraction from such noisy text is not yet
	able to provide facts with sufficiently high accuracy~\cite{DBLP:conf/ijcnlp/HanGLPYXLLZS20}.
	Another important issue where SEs can support KGs is how to cope with negation. As KGs follow the principle of the Open World Assumption, statements that are not in KG can either be false or merely missing.
	As an example, a KG may state that Manuel Blum won the Turing Award but not list him among ACM Fellows. However, we do not know whether the KG has a complete list of ACM Fellows -- such that Blum is not a Fellow -- or Blum is just missing from what is an incomplete list.
	Retrieving and analyzing webpages that provide tangible support or counter-evidence would be valuable to resolve the doubt.
	
	Referring back to Tables~\ref{tbl:proscons}~\&~\ref{tbl:needs}, further research along these lines would help improve the freshness, coverage and completeness of KGs, but there are challenges relating to how to extract accurate information relevant for the KG from potentially noisy SE results.
	
	\subsection{KG + LLM + SE}
	
	\subsubsection*{Augmentation $\rightarrow$ Ensemble $\rightarrow$ 
		Federation $\rightarrow$ Amalgamation} 
	To combine all three technologies, we envision research following a natural progression through four phases. 
	
	The \textit{augmentation} phase considers 
	a primary technology that is augmented by the other two. These are natural extensions of the research topics already discussed.
	For example, in the context of \textit{SE for KG: Knowledge Refinement}, one could  
	pursue the idea of
	extending this to \textit{SE + LLM for KG} whereby LLMs are used to help extract knowledge from sources judiciously retrieved by the SE in order to extend or refine the KG.
	Likewise, \textit{LLM for KG: Knowledge Generation} could be extended into \textit{LLM + SE for KG} by using RAG for knowledge generation.
	These two directions could actually unify the broader theme of \textit{knowledge acquisition and generation}.
	
	Moving away from augmenting a main technology, one can consider an \textit{ensemble} approach, where KGs, LLMs and SEs are peers in the information infrastructure, and a user query is delegated to the technology best adapted to address the particular type of information need. 
	Such an ensemble would likely have a natural language interface powered by an LLM, but underneath it could call KGs, LLMs or SEs. 
	Referring to Table~\ref{tbl:needs}, for example, multi-hop queries might be delegated to a KG, commonsense queries to an LLM, while queries about news events might be sent to an SE. 
	The choice would be automated, and could also address non-functional dimensions, for example, boosting efficiency by shipping simple queries to low-cost components (e.g., fact lookups on a KG).
	
	Next, we can consider the
	\textit{federation}
	of all three technologies, where, like with ensembles, user requests are routed to the best suited component, but additionally, all components can identify sub-tasks that are again shipped to other components as necessary.
	Some of these {\em delegation} approaches are already being pursued, by allowing LLMs to make API calls to external tools like calendar, calculator etc.
	(see, e.g., \cite{DBLP:conf/nips/SchickDDRLHZCS23,DBLP:journals/corr/abs-2305-15334}).
	KGs can also be part of such an ecosystem, to be invoked for fact-centric sub-tasks by an LLM or a SE.
	Putting this federation paradigm into practice still requires more research. One of the key issues is learning when to make which API calls, how to infer the values for the calls' input parameters, and how to digest the output parameters, casting them into a format suitable for the calling technology.
	When efficiency matters, additional optimization of ``execution plans'' are called for.
	Another source of structured knowledge that can be used by such a federated system is a library of functions (such as Wikifunctions~\cite{DBLP:journals/cacm/Vrandecic21}) which can be used to compute the answers to a question, or the parts of an answer that can be used to assemble a complex answer.
	
	Finally, the fourth phase, which we call \textit{amalgamation}, involves combining these technologies at a fundamental level. This might involve, for example, creating a combined neural representation of natural language and (factual) structured knowledge (with semantics), complemented with inverted indexes for search, database indexes and semantics for knowledge, and structured query languages for resolving complex factual requests. Key to this would be using consistent unambiguous tokens, for example, to align the tokens used to generate the text ``Turing award'' (per an LLM), to be able to query for those who have won such awards born in a certain region (per a KG), to be able to search for sources mentioning such winners (per an SE), etc. Such tokens could be further ``enriched'', for example, to associate them with datatypes, or to incorporate structured, semantic descriptions of tokens and how they interrelate. Another idea would be to explore ``dual neural knowledge'' whereby LLMs and KGs are applied for a dual encoding of popular entities, whereas KGs and SEs are used for long-tail information. Whatever the particular research direction, in this phase, the lines between SEs, KGs and LLMs become increasingly blurred, leading to an amalgam technology that aims to surpass the sum of its parts.
	
	The goal of research on combining these three technologies should be to emphasize the pros of the individual technologies, while suppressing (insofar as possible) their cons: referring to Table~\ref{tbl:proscons}, this would involve capturing as many +'s as possible while avoiding as many $-$'s as possible. Such a combination would then naturally lead to much better performance for factual, explanatory planning and advice questions, as covered in Table~\ref{tbl:needs}.
	
	\section{Conclusions}
	
	We have discussed Search Engines (SEs), Knowledge Graphs (KGs) and Large Language Models (LLMs) from the 
	perspective of users seeking information.
	Our analysis reveals that rather than being competitors, 
	the three technologies are indeed complementary. 
	
	SEs offer the broadest and freshest coverage and can provide precise results with tangible evidence, sometimes even for the long tail. However, SEs by themselves are inherently unable to integrate or synthesize results from multiple sources. 
	KGs can reason over multitudinous facts and synthesize results derived from multiple sources in a precise, transparent and deterministic manner.
	However, due to their quality standards and curation process, they have generally 
	weaker coverage and freshness, and often lack the nuance and diversity captured by natural language collections. 
	LLMs fall between SEs and KGs in terms of coverage and freshness,
	and can also synthesize nuanced results from multiple sources.
	However, they tend to 
	perform poorly on long-tail topics, and are prone to imprecise or invalid outputs, lack of transparency and non-determinism.
	
	Regarding users' information needs, KGs excel on complex factual queries, but do not cope well with non-factual categories.
	SEs provide support for factual and non-factual categories, but are effective only on simple queries and are inconvenient for questions whose answers do not lie in a single document; similar limitations arise also when structured data needs to be aggregated. LLMs also partially cover both factual and non-factual needs, but 
	are prone to hallucinations and bias, 
	have no formal 
	operators for analytical queries, 
	and can be trapped by false premises in questions.
	
	We argue herein that research on these technologies would benefit from keeping in mind the full diversity of the information needs of users. Although research on each of the three technologies will continue to be fruitful in the coming years along these lines, we believe that research on their combination and integration -- with users in mind -- will be particularly fruitful.
	
	\begin{acks}
		We thank the organizers of the Knowledge Graph Conference, the SEMANTICS Conference, the International Semantic Web Conference and the ACM SIGMOD Conference for inviting us to give keynotes on this topic. Hogan was supported by ANID -- Millennium Science Initiative Program -- Code ICN17\_002, and by FONDECYT Grant 1221926.
	\end{acks}
	
	\bibliographystyle{ACM-Reference-Format}
	\bibliography{refs}
	
	\appendix
	
	\section{Background}\label{sec:background}
	
	This section provides background on the three core technologies of Large Language Models, Search Engines and Knowledge Graphs.
	
	\subsection{Large Language Models}
	
	LLMs capture contextual probabilities of tokens in the parameters of a large neural network, often following the Transformer architecture \cite{DBLP:conf/nips/VaswaniSPUJGKP17}.
	The model parameters are computed by two stages of training: {\em unsupervised pre-training} and {\em supervised fine-tuning}.
	LLMs can also benefit from inference-time (i.e., post-training) techniques, most notably, {\em prompt engineering} \cite{DBLP:journals/csur/LiuYFJHN23} and {\em in-context learning} \cite{DBLP:journals/corr/abs-2301-00234}.
	
	\parai{Pre-training} is the backbone of all LLMs. To this end, the models are provided with huge corpora of human-written texts, covering Wikipedia, news articles, social media, book collections, and much more -- all at massive scale with trillions of tokens (see, e.g., \cite{DBLP:journals/corr/abs-2407-21783}).
	The usual training objective is to predict the next token in a text sequence, repeatedly in an auto-regressive manner. As the original text is available, the ground-truth is known and this entire training process is completely unsupervised (or self-supervised, as it is sometimes phrased).
	Multimodal models additionally leverage huge collections of speech, images and videos, often in combination with transcribed text and user annotations or comments. These are out of scope for this article, though.
	
	\parai{Fine-tuning} adopts a pre-trained LLM as a foundational model and adapts it for a suite of specific tasks (such as question answering, chatbot dialog, summarization etc.) by adapting its internal parameters. 
	\textit{Supervised fine-tuning} (\textit{SFT}) provides the model with labeled examples of desired input--output pairs specific to the task. A common SFT technique is to apply \textit{instruction fine-tuning} 
	\cite{DBLP:journals/corr/abs-2308-10792}, whereby the inputs include task-specific natural-language instructions paired with a desired output.
	Unlike pre-training, this stage critically depends on labeled data, typically with human annotators/contributors in the loop. Another related technique is to apply \textit{reinforcement learning from human feedback} (\textit{RLHF}), whereby a set of ranked preferences collected from human users over alternative outputs to the same model input is used to train a separate reward model to predict such preferences; the language model is then fine-tuned by the reward model to better fit its predicted human preferences. A more recent approach, called \textit{direct preference optimization} (\textit{DPO})~\cite{RafailovSMMEF23}, obviates the need for a reward model, rather embedding a loss function directly into the language model to fine-tune it for human preferences.
	In summary, an ideal LLM acquires both language skills and background knowledge from pre-training, and the ability to follow user instructions and tackle a wide variety of specific tasks from fine-tuning.
	
	\parai{In-context learning} enables the model to learn from the user's input at inference time. A key technique is \textit{prompt engineering}~\cite{DBLP:journals/csur/LiuYFJHN23}: phrasing the prompt(s) issued to the model in order to elicit a more favorable response. 
	This can take the form of giving background context about the user or the task at hand, or giving specific instructions to the system.
	A common technique is to include examples of desired inputs and outputs in the prompt, enabling \textit{few-shot learning}~\cite{DBLP:journals/corr/abs-2301-00234}. 
	Another technique, called \textit{chain-of-thought}~\cite{DBLP:conf/nips/Wei0SBIXCLZ22}, breaks down a complex request into a sequence of intermediate steps.
	In some use-cases, these techniques lead to impressive behavior, but they are also brittle in the sense that slightly different prompts can lead to inconsistent outputs.

	\parai{Retrieval-augmented generation} combines LLMs with search engines or other information-retrieval technologies to enable the ingestion of fresh Web contents for in-context learning, thus helping to mitigate the effects of stale and hallucinated outputs~\cite{DBLP:conf/nips/LewisPPPKGKLYR020,DBLP:conf/kdd/FanDNWLYCL24}. This topic is covered in more detail in the body of the article.

	\subsection{Search Engines}
	
	Search Engines (SEs) help users to retrieve relevant information from potentially massive corpora, like the Web, including news feeds, social media streams and even structured datasets such as JSON files or CSV tables.
	SEs are powered by Information Retrieval (IR) techniques~\cite{DBLP:books/aw/Baeza-YatesR2011}, classically geared towards
	matching keywords in documents, but the last decade has brought many innovations~\cite{DBLP:journals/sigir/MetzlerTBN21}, improving user experience in finding answers.
	SE technology comprises the following components
	\cite{DBLP:books/sp/23/L2023}:
	
	\parai{Collection} involves the acquisition and continuous updating of a corpus of 
	web-pages over which users can search.
	This is largely based on \textit{crawling} the Web, but also involves subscribing to feeds (e.g., news, social media) and
	importing information from databases (e.g., product catalogs).
	These sources must be revisited intermittently in order to check for updates, with the frequency dependent on the importance and rate of change of the particular page.
	Methods for spam control can block or remove malicious contents.
	The collection process works near-real-time; for example, a new post on a social network will be known to the SE within minutes.
	
	\parai{Indexing} organizes tokens, words and phrases of documents into \textit{inverted-index lists}: highly optimized data structures that enable efficiently retrieving documents that match a search request. In addition to indexing surface tokens, the SE also prepares for similarity search, aka
	soft matching, by computing an embedding vector for each document (or part of a document). This is done by running the text through a lightweight LLM such as BERT \cite{GoogleBERT}, possibly restricted to popular web sources if cost is of concern.

	\parai{Query understanding} is a key component of modern SEs, which boast a sophisticated suite of techniques for interpreting and enriching users' queries \cite{chang2020query}, ideally inferring the user intent rather than staying at the string-matching level. This includes techniques for auto-completion, suggestions for reformulations and topically related queries (e.g., based on query-and-click logs),
	contextualization with user history and other situative context
	(e.g., location, time of day, etc.)~\cite{DBLP:books/sp/23/L2023}.
	The query and the relevant context can also be fed through a 
	lightweight LLM, with the resulting embedding vector used
	as a search-intent representation for matching and ranking.
	
	\parai{Matching} involves locating occurrences of the user's keywords and short phrases as terms in the index, and retrieving all exact matches or possibly partial matches for subsequent ranking.
	When users pose full-fledged questions or give other kinds of longer inputs, this matching paradigm does not make sense.
	Instead, the query embedding vector (see above)
	can be compared, via approximate nearest-neighbor search, against
	the precomputed embeddings of the documents \cite{DBLP:journals/ftir/FanXCCMLZG22}. This LLM-based encoding can  be important as it captures the ordering and/or proximity of the query words.
	Yet another variant for matching is when the SE recognizes entity names in the query and can map some of these onto a KG with high confidence. In this case, excerpts from the KG are fetched 
	and later shown as a knowledge panel or combined with other results
	\cite{GoogleKG}.
	
	\parai{Ranking} scores matching documents in terms of their likelihood to be of interest to the user. This is based on 
	various factors such as relevance to the search (as determined during the matching phase), prominence, recency, click-through rates, and others.
	Modern SEs will often use machine learning models to combine different features into a final ranking of documents presented to users.
	
	\parai{Result presentation} involves the layout of relevant documents, and other related information, retrieved by the engine. The classical SE result page (SERP) is an ordered list of ten blue links, each with a URL and a brief preview snippet. As leading SEs often expand a user query into several vertical-domain searches, including product or event search, sponsored ads and also KG excerpts (so-called knowledge panels),
	the modern presentation is more elaborate and faceted.
	Moreover, whenever possible and particularly on limited displays like mobile phones, the SE aims to return \text{direct answers}: short text that gives a single answer, as if responding in question-answering mode. 
	Recently, SEs have started to integrate generative AI for better user experience, such as providing overviews on search topics, comparisons,
	personalized advice, and more \cite{GoogleGenAI}.
	
	\subsection{Knowledge Graphs}
	
	\newcommand{\dtype}[2]{\begin{tabular}{@{}c@{}}
		\\[-3.6ex] 
		{\sf #1}\\[-1ex] 
		{\sc \scriptsize #2}\\[-1ex] 
	\end{tabular}}

	\begin{figure}
		\begin{tikzpicture}
			\node[iri] (mb) {\sf Manuel Blum};
			
			\node[iri,right=1.2cm of mb] (tanf) {\sf TA1995}
			edge[arrin] node[above] {\sf won} (mb);
			
			\node[iri,right=1.5cm of tanf] (ta) {\sf Turing Award}
			edge[arrin] node[above] {\sf type} (tanf);
			
			\node[lit,above=0.3cm of ta] (nf) {\dtype{1995}{year}}
			edge[arrin,bend right=15] node[above] {\sf year} (tanf);
			
			\node[lit,below=0.3cm of ta] (acc) {\dtype{true}{boolean}}
			edge[arrin,bend left=15] node[below] {\sf accepted} (tanf);
			
			\node[iri,below=1.5cm of mb] (car) {\sf Caracas}
			edge[arrin] node[auto] {\sf born in} (mb);
			
			\node[iri,right=1.2cm of car] (ven) {\sf Venezuela}
			edge[arrin] node[above] {\sf part of} (car)
			edge[arrin] node[right,xshift=0.1cm] {\sf citizenship} (mb);
			
			\node[iri,right=1.2cm of ven] (la) {\sf Latin America}
			edge[arrin] node[above] {\sf part of} (ven);
		\end{tikzpicture}
		\caption{Knowledge graph describing Manuel Blum \label{fig:kg}}
		
		\Description[Knowledge graph describing Manuel Blum]{Graphical depiction of knowledge graph describing Manuel Blum, with nodes representing him, the Turing Award, the year 1995, and labeled edges connecting these nodes.}
	\end{figure}
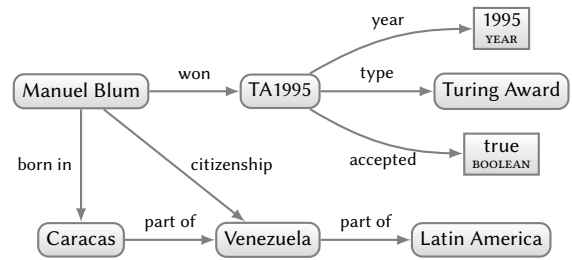
	
	The modern notion of Knowledge Graphs (KGs) arose in the context of improving search engines~\cite{DBLP:journals/cacm/GutierrezS21}. 
	In 2012, the Google Knowledge Graph was proposed to boost search by ``\textit{things, not strings}''~\cite{GoogleKG}:
	when matching a string like ``\texttt{Manuel Blum}'' against web contents, the SE can recognize that the string likely references a real-world entity.
	This requires a structured repository of real-world entities, 
	their types (e.g., computer scientist), their attributes (e.g., birth date), and the relations between entities (e.g., Blum has birthplace Caracas).
	Figure~\ref{fig:kg} shows a toy KG where nodes represent entities (rounded corners) or types
	(square corners), and edges represent the relations between them.  KGs enable or enhance a variety of 
	foundational functionalities including the following:
	
	\parai{Construction} involves the initial creation of a KG from base sources of data. Initially this can involve a mix of information extraction (for text and other unstructured sources) and mapping (for structured sources) techniques. Key to constructing a high-quality KG is the disambiguation process, which assigns unique identifiers to distinct entities and relation types in the presence of synonyms (same string, different entities) and homonyms (different strings, same entity). For example, the KG distinguishes \textit{Caracas} the city (\texttt{Q1533} in Wikidata), and \textit{Caracas} the album by Lou Donaldson (\texttt{Q5037560} in Wikidata). 
	Together with its knowledge of attribute values, relations, alias names and more, this makes the KG a key asset in text-mining tasks with \textit{entity linking} -- mapping ambiguous names onto unique entities \cite{DBLP:journals/tkde/ShenLLHWY23} -- which in turn is used to help construct and enrich KGs from text.
	
	\balance
	
	\parai{Completion} aims to ``fill in the gaps'' inevitably left in KGs by integrating data from diverse sources. We consider two paradigms for KG completion: \textit{deductive} and \textit{inductive}. Deductive reasoning enables automated inference of new edges in the knowledge graph when complemented by rules or ontologies.
	For example, we may declare that the \textsf{part-of} relation is \textit{transitive}, such that from Figure~\ref{fig:kg} we can infer a new edge indicating that {\sf Caracas} is {\sf part of} {\sf Latin America}. We can further assert the \textit{property chain} ``$\textsf{born in} \circ \textsf{part of} \sqsubseteq \textsf{born in}$'', inferring two new edges indicating that {\sf Manuel Blum} was {\sf born in} {\sf Venezuela} and {\sf Latin America}. Assuming the base graph and the axioms encoded by the rules/ontologies to be correct, such inferences will be correct. On the other hand, \textit{inductive}
	reasoning does not require the specification of rules/ontologies, but rather learns abstract patterns from the KG and applies these patterns for completion. \textit{Knowledge graph embeddings}~\cite{DBLP:journals/tkde/WangMWG17} learn tensor representations for nodes and edges in the graph that can be used for \textit{link prediction}, e.g., to predict \textsf{citizenship}s based on other available information in the knowledge graph (e.g., \textsf{born in} relations). \textit{Graph neural networks}~\cite{DBLP:journals/tnn/WuPCLZY21} can be used to classify nodes, edges or graphs; for example, given a list of Turing Award laureates, such networks can be used to predict future winners. More recently, hybrid approaches that blend these deductive and inductive paradigms -- which fall into the broader \textit{neurosymbolic} category -- have been proposed and explored in the literature~\cite{DBLP:journals/tgdk/PanRKSCDJO0LBMB23}.
	
	\parai{Refinement} involves the detection and resolution of issues affecting a KG's \textit{quality} (i.e., fitness for purpose). While similar methods as used for completion can be adapted for refinement -- e.g., rules/ontologies can be used to identify logical inconsistencies in a KG, while embeddings can be used to identify edges with low \textit{plausibility} -- dedicated methods also exist for refinement. A key such technique is to leverage \textit{shapes}, which encode constraints applicable to the KG that can validate and improve its quality~\cite{DBLP:series/synthesis/2017Gayo}. A shape might define that nodes of \textsf{type} \textsf{Turing Award} must specify a value for \textsf{year} (passed by {\sf TA1995} in Figure~\ref{fig:kg}), or that all winners of a \textsf{Turing Award} must have the occupation \textsf{Computer Scientist} (failed by {\sf Manuel Blum} in Figure~\ref{fig:kg}).
	
	\parai{Search \& querying} involves extracting information from the KG relevant for a particular user or a particular task. KGs often contain text, which can be used to leverage search in a manner similar to SEs. However, the structure of KGs further permits evaluating database-style queries expressed in graph query languages, which include many features from relational databases (joins, aggregation, etc.) as well as querying arbitrary-length paths~\cite{DBLP:journals/csur/AnglesABHRV17}; for example, a \textit{regular path query} such as:
	\begin{center} \resizebox{!}{2.3ex}{\begin{tikzpicture}[baseline=-0.75ex] 	\node[iri,dashed] (ni) {\sf ?w};
				\node[iri,right=2.4cm of ni] (oo) {\sf Latin America} edge[arrin] node[lab] {\sf born in/part of{\LARGE *}} (ni);
		\end{tikzpicture}} 
	\end{center}
	\noindent
	will match paths of one ``\textsf{born in}'' edge followed by zero or many (indicated by ``{\Large *}'') ``\textsf{part of}`'' edges, returning {\sf Manuel Blum} from Figure~\ref{fig:kg} as a result for the variable {\sf ?w}.
	Queries over the structured KG provide {\em complete answer sets} with respect to that KG. For example, if the KG contains all ACM Fellows with their countries, a simple group-by query can return a list of countries or continents with the respective numbers of fellows. One challenge is that writing structured queries can be difficult for many users. \textit{Knowledge graph question answering} (\textit{KGQA}) is a task that focuses on resolving natural language questions over the KG, often by applying semantic parsing to convert the user question into a structured query.
	
\end{document}